\documentclass{article}

\PassOptionsToPackage{numbers, compress}{natbib}
\usepackage[nonatbib,preprint]{neurips_2025}
\usepackage{math-macros}

\usepackage[utf8]{inputenc} 
\usepackage[T1]{fontenc}    
\usepackage[colorlinks=true,allcolors=black]{hyperref}       
\usepackage{url}            
\usepackage{booktabs}       
\usepackage{amsfonts}       
\usepackage{nicefrac}       
\usepackage{microtype}      
\usepackage{xcolor}         
\usepackage{amsthm}
\usepackage{enumitem}

\usepackage[textsize=scriptsize]{todonotes}
\usepackage{algorithm}
\usepackage{algpseudocode}
\usepackage[style=nature,doi=false,isbn=false,url=false,date=year]{biblatex}
\usepackage{enumitem}

\AtEveryBibitem{%
  \clearlist{language}%
  \clearname{editor}%
  \clearlist{publisher}%
  \clearfield{pages}%
  \clearfield{pagetotal}%
  \clearfield{location}%
  \clearfield{edition}%
  \clearfield{series}%
}

\graphicspath{graphics}
\bibliography{references}

\title{Hierarchical mixtures of Gaussians for joint dimensionality reduction and clustering}

\author{%
  Sacha Sokoloski\\
  University of Tübingen\\
  \texttt{sacha.sokoloski@mailbox.org} \\
  \And%
  Philipp Berens \\
  University of Tübingen\\
  \texttt{philipp.berens@uni-tuebingen.de} \\
}

\theoremstyle{plain}

\begin{document}

\maketitle

\begin{abstract}

	We introduce hierarchical mixtures of Gaussians (HMoGs), which unify dimensionality reduction and clustering into a single probabilistic model. HMoGs provide closed-form expressions for the model likelihood, exact inference over latent states and cluster membership, and exact algorithms for maximum-likelihood optimization. The novel exponential family parameterization of HMoGs greatly reduces their computational complexity relative to similar model-based methods, allowing them to efficiently model hundreds of latent dimensions, and thereby capture additional structure in high-dimensional data. We demonstrate HMoGs on synthetic experiments and MNIST, and show how joint optimization of dimensionality reduction and clustering facilitates increased model performance. We also explore how sparsity-constrained dimensionality reduction can further improve clustering performance while encouraging interpretability. By bridging classical statistical modelling with the scale of modern data and compute, HMoGs offer a practical approach to high-dimensional clustering that preserves statistical rigour, interpretability, and uncertainty quantification that is often missing from embedding-based, variational, and self-supervised methods.

\end{abstract}

\section{Introduction}

Clustering --- the grouping of data based on similarity --- is an essential part of knowledge discovery across scientific domains. The growing complexity of modern datasets thus poses a challenge, as many clustering algorithms suffer from the so-called ``curse of dimensionality'', and exhibit limited performance when modelling high-dimensional data. This arises not only because model complexity may scale poorly with dimension, but also because the similarity metrics used to evaluate model performance tend to break down in high-dimensional spaces~\cite{beyer_when_1999,assent_clustering_2012}.

A common approach to this challenge is to first project the high-dimensional data into a lower-dimensional latent space that captures its essential structure, and then cluster the projected data within this latent space. Such two-stage approaches are widely applied to problems in image processing~\cite{houdard_high-dimensional_2018,minaee_image_2022} and time-series analysis~\cite{liao_clustering_2005,paparrizos_bridging_2024}, and in fields ranging from neuroscience~\cite{lewicki_review_1998,baden_functional_2016} to bioinformatics~\cite{witten_framework_2010,duo_systematic_2020}. Nevertheless, two-stage algorithms separate the optimization of dimensionality reduction from clustering, and may thereby discard information that is critical for cluster identification. For example, when using PCA, the directions of maximum variance might not be the directions that best separate clusters~\cite{chang_using_1983,mclachlan_finite_2019}. Meanwhile, modern approaches like deep clustering and self-supervised methods can capture nonlinear latent features~\cite{xie_unsupervised_2016,chen_simple_2020,grill_bootstrap_2020}, but often sacrifice statistical guarantees and probabilistic interpretability, making principled model selection and uncertainty quantification difficult.

To address these limitations we consider a family of two-stage models that implement dimensionality reduction with a linear Gaussian model (LGM) such as PCA or factor analysis (FA), and clustering with a mixture of Gaussians (MoG). By applying a novel theory of analytically tractable, latent variable exponential family models~\cite{sokoloski_unified_2024}, we show how these two stages can be combined and generalized into a single probabilistic model that we call a hierarchical mixture of Gaussians (HMoG). With HMoGs we make the following novel contributions in the space high-dimensional clustering:
\begin{enumerate}[nosep, leftmargin=2em]
	\item \textbf{Unified exponential family framework:} We develop a hierarchical, latent variable, exponential family model of dimensionality reduction and clustering, that unifies existing analysis pipelines.
	\item \textbf{Scalable implementation:} We develop a novel parameterization for HMoGs that greatly reduces its computational complexity, allowing it to scale to large samples of high-dimensional data.
	\item \textbf{Exact algorithms:} We show how HMoGs avoid approximations and assumptions common to most other frameworks, and afford exact expressions and scalable algorithms for
	      \begin{enumerate}[nosep]
		      \item latent-variable inference and soft cluster-assignment,
		      \item maximum-likelihood learning for joint dimensionality reduction and clustering,
		      \item probabilistic cluster merging based on cluster uncertainty, and
		      \item model evaluation, and thereby principled model selection.
	      \end{enumerate}
\end{enumerate}

We provide an implementation of HMoGs using the JAX numerics library, and use it to demonstrate the effectiveness of HMoGs on MNIST~\cite{lecun_gradient-based_1998}. We also explore the value of adding $\ell_1$ sparsity to the dimensionality reduction, and show how HMoGs learn center-surround features --- akin to sparse-codes observed in primary visual cortex~\cite{olshausen_emergence_1996,rehn_network_2007} --- that further boosts clustering accuracy.

\subsection{Related Work}

Dimensionality reduction and clustering is a methodologically heterogeneous field, but we may break it down along the following lines:

\begin{description}[nosep, leftmargin=0em]
	\item[Classical two-stage approaches:] Pipelines that sequentially apply classic dimensionality reduction (typically PCA or factor analysis) followed by clustering techniques (e.g., k-means) remain widely used due to their simplicity, speed, and interpretability (\cite{witten_framework_2010,duo_systematic_2020}). Nevertheless, these approaches optimize separate objectives for dimensionality reduction and clustering.
	\item[Model-based clustering:] Our work is most directly related to previous methods on mixtures of PCA/FA~\cite{ghahramani_em_1996,tipping_mixtures_1999,houdard_high-dimensional_2018,gormley_model-based_2023}. In contrast with these methods, however, our model has an exponential family expression that affords a more efficient parameterization, allowing it to scale to modern datasets.
	\item[Nonlinear two-stage methods:] Nonlinear dimensionality reduction techniques such as t-SNE and UMAP~\cite{maaten_visualizing_2008,mcinnes_umap_2020} can identify complex clusters in data when combined with clustering algorithms like DBSCAN or HDBSCAN~\cite{ester_density-based_1996,campello_hierarchical_2015}. Methods in self-supervised and contrastive learning have also been used to learn representations that are deployed for downstream clustering~\cite{chen_simple_2020,grill_bootstrap_2020}. These methods excel at clustering complex data yet suffer the same limitations as classic two-stage methods, while also sacrificing a degree of interpretability.
	\item[End-to-end deep clustering:] Recent work has seen the development of end-to-end deep learning architectures that simultaneously learn representations and clusterings. These include Deep Embedded Clustering (DEC) and Joint Unsupervised Learning (JULE), which integrate autoencoders with clustering objectives~\cite{xie_unsupervised_2016,yang_joint_2016,ji_invariant_2019}. While these approaches achieve high-performance, they typically sacrifice the probabilistic foundations needed for principled model selection and uncertainty quantification.
\end{description}

\section{Background}

In this paper we presume familiarity with LGMs, mixture models, and exponential family theory --- for a thorough treatment of LGMs see~\cite{bishop_pattern_2006} and of exponential families see~\cite{wainwright_graphical_2008}. We also rely on recent recent work that unifies LGMs and mixture models under the theory of conjugated, exponential family harmoniums~\cite{sokoloski_unified_2024}. In this section we provide a brief overview and develop key notation.

\subsection{Linear Gaussian models and mixture models as latent variable models}\label{sec:lvms}

LGMs and MoGs provide probabilistic approaches to dimensionality reduction and clustering, respectively. To see this, let us suppose we have $n$ observations $\V x^{(1)}, \dots, \V x^{(n)}$ of the $d_X$-dimensional random variable $X$, and that $X$ is influenced by some $d_Y$-dimensional latent variable $Y$. In the maximum-likelihood framework, we fit a statistical model $p(\V x;\eprms)$ by maximizing the log-likelihood $\frac{1}{n}\sum_{i=1}^{n}\log p(\V x^{(i)};\eprms)$ given the data with respect to the parameters $\eprms$. We may extend this framework to an LVM $p(\V x,\V y;\eprms)$ by maximizing the log-likelihood of the marginal $p(\V x;\eprms)$ of $p(\V x,\V y;\eprms)$.

Now an LGM $p(\V x, \V y)$ is simply a $d_X+d_Y$ dimensional multivariate normal (MVN) distribution, where $X$ and $Y$ are continuous variables of dimensions $d_X$ and $d_Y$, respectively. As a consequence, the marginal distributions $p(\V x)$ and $p(\V y)$, as well as the conditional distributions $p(\V x \mid \V y)$ and $p(\V y \mid \V x)$ are also MVNs~\cite[see][]{bishop_pattern_2006}. Probabilistic PCA (PPCA) and FA are forms of LGM that find latent structure through a ``loading matrix'' $\V W$. They have the form $p(\V x \mid \V y; \V \mu, \V \Sigma) = \Normal(\V \mu + \V W \cdot \V y, \V \Sigma)$ and $p(\V y) = \Normal(\V 0, \V I)$, where PPCA has an isotropic covariance matrix $\V \Sigma = \sigma \V I$, and FA a diagonal covariance matrix $\V \Sigma = \diag(\V \Psi)$ (``classic'' PCA is recovered in the limit as $\sigma \to 0$~\cite{roweis_em_1997}). In both cases, dimensionality reduction from a datapoint $\V x^{(i)}$ to its projection $\V y^{(i)}$ is given by the mean of the posterior distribution $p(\V y \mid \V x)$ such that $\V y^{(i)} = \mathbb E[Y \mid X = \V x] = \V W^{\V T} \cdot {(\V W \cdot \V W^{\V T} + \V \Sigma)}^{-1}\cdot \V x$.

A mixture model is a latent variable model $p(\V y, k)$ over observables $Y$, where the latent variable $K$ is an index ranging from $0$ to $d_K-1$. In the probabilistic formulation, the component distributions of the mixture model are given by the conditional distribution $p(\V y \mid k)$ for each index $k$, the weights $\V \pi$ of the mixture are the prior probabilities $\pi_i = p(k=i)$, and the mixture distribution (the weighted sum of the components) is given by the marginal distribution $p(\V y)$. MoGs, then, are simply the case where each component $p(\V y \mid k=i) = \Normal(\V \mu_i, \V \Sigma_i)$ is an MVN with mean $\V \mu_i$ and covariance $\V \Sigma_i$.

\begin{figure}[t]

	\includegraphics{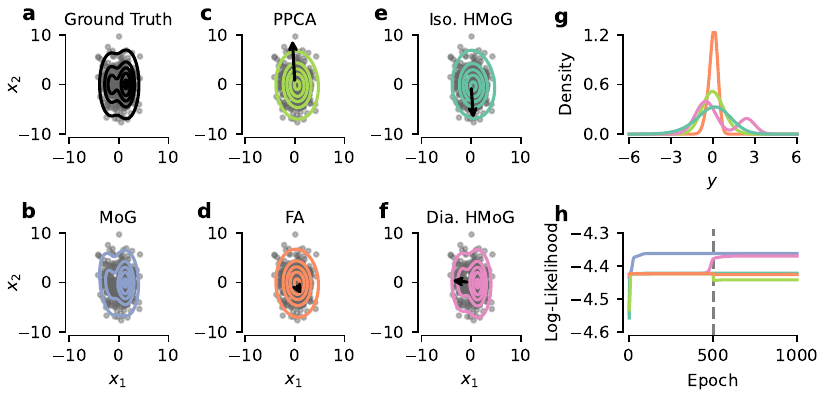}

	\caption{\textit{Limitations of two-stage models.} \textbf{a-f:} A sample (black dots) from a mixture of two Gaussians (\textbf{a}, black contours). Contours of sample fits from a MoG with two components (\textbf{b}, blue), PPCA (\textbf{c}, green) and FA (\textbf{d}, orange), an isotropic HMoG (\textbf{e}, turqoise), and a diagonal MoG (\textbf{f}, purple) each with two components. Loading vectors (black arrow) learned by PPCA (\textbf{c}), FA (\textbf{d}), and the LGM layers of the isotropic (\textbf{e}) and diagonal (\textbf{f}) HMoGs. \textbf{g:} Latent densities (lines) learned by two-stage PPCA (green), two-stage FA (orange), an isotropic HMoG (turqoise), and a diagonal HMoG (purple). \textbf{h:} Log-likelihood trajectories (lines) of all models (same colour schemes), and starting epoch of second of two-stage training (dashed line).}\label{fig:synthetic}

\end{figure}

To begin motivating our integrated hierarchical approach to dimensionality reduction and clustering, let us consider a sample from a ground-truth MoG with two components, designed so that its maximum variance is orthogonal to the direction that determines cluster membership (Fig.~\ref{fig:synthetic}\textbf{a}). LGMs and MoGs can be fit to data with EM, and unsurprisingly we can recover our ground-truth distribution by fitting the sample with a MoG (Fig.~\ref{fig:synthetic}\textbf{b}). LGMs can only represent normally-distributed data, yet may capture different features depending on their covariance structure. When, for example, we fit PPCA to the sample, its loading matrix $\V W$ points in the direction of maximum variance (Fig.~\ref{fig:synthetic}\textbf{c}), whereas FA mostly explains the data with its diagonal variances $\V \Psi$, and its loading matrix is near $\V 0$ (Fig.~\ref{fig:synthetic}\textbf{d}). Later we will see how the loading matrices learned by PPCA and FA prevent them from effectively supporting hierarchical clustering of this synthetic dataset.

\subsection{Linear Gaussian models and mixture models as exponential family harmoniums}\label{sec:harmoniums}

We next show how to interpret LGMs and mixture models as exponential families --- in particular, as classes of model known as exponential family harmoniums~\cite{smolensky_information_1986,welling_exponential_2005}. To begin, an exponential family is a statistical model defined by a sufficient statistic $\V s$ and base measure $\nu$, and has the form $p(x; \eprms) = e^{\V s(x) \cdot \eprms - \psi(\eprms)}\nu(x)$, where $\eprms$ are the natural parameters and $\psi$ is the log-partition function. In general, we denote an exponential family by the pair $(\V s, \nu)$.

There are two exponential families that are particularly relevant for our purposes. Firstly, the family of $d_X$-dimensional MVN distributions is the exponential family with base measure $\nu = {(2\pi)}^{-\frac{n}{2}}$ and sufficient statistic $\V s(\V x) = (\V x, \V x \otimes \V x)$, where $\otimes$ is the outer product operator. Consequently its log-partition function is given by
\begin{equation}\label{eq:lgm-log-partition}
	\psi(\eprms) = -\frac{1}{4} \eprms^\mu \cdot {\iprms^{\sigma}}^{-1} \cdot \eprms^\mu - \frac{1}{2}\log |-2 \iprms^{\sigma}|,
\end{equation}
where $\eprms^\mu$ are the location parameters that multiply with $\V x$, and $\iprms^\sigma$ the precision parameters that multiply with $\V x \otimes \V x$ in $\V s(\V x)$, respectively. Secondly, the family of categorical distributions over indices $0, \dots, d_K-1$ is the exponential family with base measure $\nu = 1$, and sufficient statistic given by a so-called ``one-hot'' vector, so that $\V s(0) = \V 0$, and $\V s(k)$ is a $d_K-1$ length vector with all zero elements except for a 1 at element $k-1$. The log-partition function of the categorical family is given by the ``LogSumExp'' function over $\eprms$
\begin{equation}\label{eq:categorical-log-partition}
	\psi_K(\eprms) = \log \sum_{i=0}^{d_K-1} \exp^{\theta_i}.
\end{equation}

In general, an exponential family harmonium is as a kind of product exponential family, which includes various models (such as LGMs and mixture models) as special cases~\cite{welling_exponential_2005, sokoloski_unified_2024}. Given two exponential families $(\V s_X, \nu_X)$, and $(\V s_Y, \nu_Y)$, respectively, a harmonium $(\V s_{XY}, \nu_{XY})$ is the model
\begin{equation}\label{eq:harmonium-model}
	p(x,y; \eprms_X, \eprms_Y, \iprms_{XY}) = e^{\V s_X(x) \cdot \eprms_X + \V s_Y(y) \cdot \eprms_Y + \V s_X(x) \cdot \iprms_{XY} \cdot \V s_Y(y) - \psi_{XY}(\eprms_X, \eprms_Y, \iprms_{XY})}\nu_X(x) \nu_Y(y),
\end{equation}
where $\eprms_X$, $\eprms_Y$, and $\iprms_{XY}$ are the observable biases, latent biases, and interactions, respectively. A harmonium is indeed an exponential family, with base measure $\nu_{XY}(x,y) = \nu_X(x)\nu_Y(y)$ and sufficient statistic $\V s_{XY}(x,y) = (\V s_x(x), \V s_Y(y), \V s_X(x) \otimes \V s_Y(y))$, where $\eprms_{XY} = (\eprms_X, \eprms_Y, \iprms_{XY})$ are the natural parameters, and $\psi_{XY}$ is the log-partition function.

To re-express LGMs and mixture models as harmoniums, let $(\V s_X, \nu_X)$, $(\V s_Y, \nu_Y)$, and $(\V s_K, \nu_K)$ be the $d_X$-dimensional MVN family, the $d_Y$-dimensional MVN family, and the categorical family over $d_K$ indices, respectively. On one hand, we may express an LGM $(\V s_{XY}, \nu_{XY})$ in the harmonium form $p(\V x, \V y; \eprms_X, \eprms_Y, \iprms_{XY}) \propto e^{\V s_X(\V x) \cdot \eprms_X + \V s_Y(\V y) \cdot \eprms_Y + \V x \cdot \iprms_{XY} \cdot \V y}$, where the harmonium sufficient statistic $\V s_{XY}$ is constrained so that there are no second order interactions between $\V x$ and $\V y$. In comparison with the LGM parameterizations from Section~\ref{sec:lvms}, the parameters $\eprms^\mu_X$ and $\iprms^\sigma_X$ correspond to $\V \mu$ and $\V \Sigma$, respectively, $\iprms_{XY}$ correspond to the parameters $\V W$, and $\eprms_Y$ parameterizes the prior~\cite{sokoloski_unified_2024}. On the other hand, we may express a MoG $(\V s_{YK}, \nu_{YK})$ in the form $p(\V y, k; \eprms_Y, \eprms_K, \iprms_{YK}) \propto e^{\V s_Y(\V y) \cdot \eprms_Y + \V s_K(k) \cdot \eprms_K + \V s_Y(\V y) \cdot \iprms_{YK} \cdot \V s_K(k)}$. In this case the natural parameters $\eprms_Y$ and $\iprms_{YK}$ correspond to the parameters $\V \mu_1, \dots, \V \mu_k$ and $\V \Sigma_1, \dots, \V \Sigma_k$ of the mixture components of the MoG, and the parameters $\eprms_K$ correspond to the mixture weights $\V \pi$ of the MoG~\cite{sokoloski_unified_2024}.

\subsection{Linear Gaussian models and mixture models as conjugated harmoniums}\label{sec:conjugated}

In general, the conditional distributions $p(x \mid y)$ and $p(y \mid x)$ of a harmonium are exponential family distributed, whereas the marginals $p(x)$ and $p(y)$ are not. Yet for an LGM $p(\V x, \V y)$ both its posterior $p(\V y \mid \V x)$ and prior $p(\V x)$ are MVN distributions, and for a MoG $p(\V y, k)$ both its posterior $p(k \mid \V y)$ and prior $p(k)$ are categorical distributions. A prior is referred to as ``conjugate'' when it has the same form as its posterior, and recent work has developed a theory of ``conjugated harmoniums'' for which their priors and posteriors have the same form~\cite{sokoloski_unified_2024}.

The posterior $p(y \mid x)$ of a harmonium $p(x, y; \eprms_X, \eprms_Y, \iprms_{XY})$ is given by
\begin{equation}\label{eq:posterior}
	p(y \mid x) = e^{\V s_Y(y) \cdot (\eprms_Y + \V s_X(x) \cdot \iprms_{XY}) - \psi_Y(\eprms_Y + \V s_X(x)\cdot \iprms_{XY})}\nu_Y(y),
\end{equation}
where $\psi_Y$ is the log-partition function of $(\V s_Y, \nu_Y)$. At a given $x$, $p(y \mid x)$ is thus in the exponential family $(\V s_Y,\nu_Y)$ with parameters $\eprms_X + \iprms_{XY} \cdot \V s_Y(y)$, and we can reduce the question of whether a harmonium prior is conjugate to whether the prior is also in the exponential family $(\V s_Y,\nu_Y)$. Where $\psi_X$ is the log-partition function of $(\V s_X, \nu_X)$, it can be shown that if
\begin{equation}\label{eq:conjugation}
	\psi_X(\eprms_X + \iprms_{XY} \cdot \V s_Y(y) ) = \V s_Y(y) \cdot \rprms_Y(\eprms_X, \iprms_{XY}) + \psi_X(\eprms_X),
\end{equation}
for some parameters $\rprms_Y(\eprms_X, \iprms_{XY})$, then $p(x)$ is in $(\V s_Y, \nu_Y)$ with parameters $\eprms_Y + \rprms_Y(\eprms_X, \iprms_{XY})$. When Eq.~\ref{eq:conjugation} is satisfied, we say that the harmonium $p(x, y)$ is conjugated, and we refer to $\rprms_Y(\eprms_X, \iprms_{XY})$ as the conjugation parameters.

Equation~\ref{eq:conjugation} is a strong constraint, yet both LGMs and mixture models are indeed conjugated harmoniums. In particular, for an LGM $p(\V x, \V y; \eprms_X, \eprms_Y, \iprms_{XY})$, the conjugation parameters corresponding to $\eprms_Y^\mu$ and $\iprms^\sigma_Y$ are
\begin{align}
	\rprms_Y^\mu    & = -\frac{1}{2} \iprms_{YX} \cdot {\iprms_X^{\sigma}}^{-1} \cdot \eprms^\mu_X,\label{eq:lgm-conjugation-mean} \\
	\V P_Y^{\sigma} & = -\frac{1}{4} \iprms_{YX} \cdot {\iprms_X^{\sigma}}^{-1} \cdot \iprms_{XY},\label{eq:lgm-conjugation-cov}
\end{align}
where $\iprms_{YX}$ is the transpose of $\iprms_{XY}$. On the other hand, the conjugation parameters $\rprms_K$ of a mixture model $p(\V y, k; \eprms_Y, \eprms_K, \iprms_{YK})$  are
\begin{equation}\label{eq:mixture-conjugation}
	\rho_{K,i} = \psi_Y(\eprms_Y + \eprms_{YK,i}) - \psi_Y(\eprms_Y),
\end{equation}
where $\eprms_{YK,i}$ is the $i$th column of $\iprms_{YK}$.

The conjugation parameters facilitate numerous computations and algorithms for conjugated harmoniums. In particular, the log-partition function of a conjugated harmonium $p(x,y)$ is given by
\begin{equation}\label{eq:conjugated-log-partition}
	\psi_{XY}(\eprms_X, \eprms_Y, \iprms_{XY}) = \psi_X(\eprms_X) + \psi_Y(\eprms_Y + \rprms_Y(\eprms_X, \iprms_{XY})).
\end{equation}
This allows us to express the marginal of the model over observables by
\begin{equation}\label{eq:harmonium-marginal}
	p(x) = e^{\V s_X(x) \cdot \eprms_X - \psi_X(\eprms_X) + \psi_Y(\eprms_Y + \V s_X(x) \cdot \iprms_{XY}) - \psi_Y(\eprms_Y + \rprms_Y(\eprms_X, \iprms_{XY}))}\nu_X (x).
\end{equation}
The log-partition function $\psi_{XY}$ of a harmonium is typically an intractable object, yet for a conjugated harmonium it is tractable up to the log-partition functions $\psi_X$ and $\psi_Y$. For PPCA and FA, in particular, the observable precisions $\iprms^\sigma_X$ are constrained to be isotropic and diagonal matrices, respectively, permitting tractable computation of $\psi_X(\eprms_X)$ (Eq.~\ref{eq:lgm-log-partition}) even when the dimension $d_X$ is large.

\section{Theory}

Here we develop the theory of HMoGs, which are hierarchical models of high-dimensional observations $X$, lower-dimensional latent representations $Y$, and cluster-indices $K$. Intuitively, an HMoG comprises a dimensionality reduction layer given by an LGM such as PPCA or FA, and a clustering layer given by a MoG, yet where the layers are jointly optimized to better resolve clusters in the observations. We formally show how to combine an LGM and a MoG into an HMoG in Section~\ref{sec:hmogs}, and show in particular how an HMoG is a form of hierarchical, conjugated harmonium. In Section~\ref{sec:complexity} we identify a compact parameterization of HMoGs, and show how to greatly reduce the computational complexity of inference, model evaluation, and learning.


\subsection{A hierarchical mixtures of Gaussians}\label{sec:hmogs}

We exploit the probabilistic formulation of an LGM $p(\V x, \V y)$ and a MoG $p(\V y, k)$ to define an HMoG as the hierarchical model $p(\V x, \V y, k) = p(\V x~\mid~\V y) p(\V y, k)$. Intuitively, we define an HMoG by taking a PCA or FA model, and swapping out the standard normal prior for a MoG. Formally, by putting together the exponential family expressions for $p(\V x \mid \V y)$ and $p(\V y, k)$ from Section~\ref{sec:harmoniums}, we may write the HMoG model as
\begin{multline}\label{eq:hmog-model}
	p(\V x,\V y, k; \eprms_X, \eprms_Y, \eprms_K, \iprms_{XY}, \iprms_{YK}) = e^{\eprms_X \cdot \V s_X(\V x) + \eprms_Y \cdot \V s_Y(\V y) + \eprms_K \cdot \V s_K(k)}
	\\ \cdot e^{\V x \cdot \iprms_{XY} \cdot \V y + \V s_Y(\V y) \cdot \iprms_{YK} \cdot \V s_K(k) - \psi_{XYK}(\eprms_X, \eprms_Y, \eprms_K, \iprms_{XY}, \iprms_{YK})}\nu_X(\V x)\nu_Y(\V y)\nu_K(k).
\end{multline}
By construction it follows that an HMoG is an exponential family with sufficient statistics $\V s_{XYK} = (\V s_X, \V s_Y, \V s_K, \V x \otimes \V y, \V s_Y \otimes \V s_K)$, base measure $\nu_{XYK} = \nu_X \cdot \nu_Y \cdot \nu_K$, and log-partition function $\psi_{XYK}$.

The expression for the HMoG model (Eq.~\ref{eq:hmog-model}) can also be rearranged into the expression for a harmonium model (Eq.~\ref{eq:harmonium-model}), such that an HMoG is a form of harmonium defined by the MVN family $(\V s_X, \nu_X)$ and a MoG $(\V s_{YK}, \nu_{YK})$, constrained to have no interactions between $\V s_X$ and $\V s_K$. Based on Equation~\ref{eq:posterior}, we may therefore express the posterior $p(\V y, k \mid \V x)$ of an HMoG as
\begin{equation}\label{eq:hmog-posterior}
	p(\V y, k \mid \V x) = e^{\V s_{YK}(\V y, k) \cdot (\eprms_{YK} + \V x \cdot \iprms_{XY}) - \psi_{YK}(\eprms_{YK} + \V x \cdot \iprms_{XY})}\nu_{YK}(\V y, k),
\end{equation}
where we use $\eprms_{YK} + \V x \cdot \iprms_{XY}$ as short-hand for $(\eprms_Y + \V x \cdot \iprms_{XY}, \iprms_{XY}, \eprms_K)$.

Because the posterior $p(\V y, k \mid \V x)$ of an HMoG is itself a MoG, an HMoG is a conjugated harmonium if its prior $p(\V y, k)$ is also a MoG. Because an HMoG is hierarchical and there are no interactions between $\V s_X$ and $\V s_K$, Equation~\ref{eq:conjugation} for an HMoG reduces to Equation~\ref{eq:conjugation} for an LGM\@. Consequently, an HMoG is a conjugated harmonium with the same conjugation parameters $\rprms_Y(\eprms_X, \iprms_{XY})$ as an LGM (Eqs.~\ref{eq:lgm-conjugation-mean}~\&~\ref{eq:lgm-conjugation-cov}), and $p(\V y, k)$ is a MoG with parameters $\eprms_{YK} + \rprms_Y = (\eprms_Y + \rprms_Y, \iprms_{YK}, \eprms_K)$.

Based on Equation~\ref{eq:harmonium-marginal} we may express the observable distribution of an HMoG as
\begin{equation}\label{eq:hmog-marginal}
	p(\V x) = e^{\V s_X(\V x) \cdot \eprms_X + \psi_{YK}(\eprms_{YK} + \V x \cdot \iprms_{XY}) - \psi_X(\eprms_X) - \psi_{YK}(\eprms_{YK} + \rprms_Y)}\nu_X (\V x),
\end{equation}
and thus evaluate the performance of an HMoG on data. Moreover, by taking the logarithm and (automatically) differentiating Equation~\ref{eq:hmog-marginal} we may iteratively optimize the parameters of an HMoG.

Taken together, Equations~\ref{eq:hmog-posterior} and~\ref{eq:hmog-marginal} thus provide a complete specification for inference, evaluation, and learning for HMoGs. With the basic HMoG equations in place, let us complete the synthetic data experiment we began in Section~\ref{sec:lvms}. We consider four new models: \textit{(i)} Two-Stage PPCA and \textit{(ii)} Two-Stage FA, where we fit PPCA and FA to data, and then fit a MoG to the corresponding projected data, respectively. We also consider \textit{(iii)} an ``isotropic HMoG'' with observable precision matrix $\iprms^\sigma_X = \theta_X^\sigma \V I$, as well as \textit{(iv)} a ``diagonal HMoG'' with diagonal precision matrix $\iprms^\sigma_X = \diag(\eprms^\psi)$.

We fit an isotropic and a diagonal HMoG each with two clusters to the sample from the ground truth MoG, and find that the former is unable to capture the clusters of the data (Fig.~\ref{fig:synthetic}\textbf{e}), in contrast with the latter (Fig.~\ref{fig:synthetic}\textbf{e}). To complete our two-stage model fitting, we fit a MoG to the projected data of the learned PPCA and FA models, and find that neither learns a clustered representation in the latent space (Fig.~\ref{fig:synthetic}\textbf{f}), and that the log-likelihood of the complete two-stage model in fact decreases through this ad-hoc learning procedure (Fig.~\ref{fig:synthetic}\textbf{g}). In the case of our HMoGs, we find that the isotropic HMoG fails whereas the diagonal HMoG succeeds in learning a clustered representation in the latent space (Fig.~\ref{fig:synthetic}\textbf{f}). Moreover, whereas the log-likelihood of the isotropic HMoG does not exceed that of the base LGMs, the diagonal HMoG is the only model able to match the performance of a MoG.

Our synthetic experiment suggests that the more flexible noise model of FA facilitates better clustering when integrated into an HMoG. While the precise interaction between the noise model and the MoG prior is complex, empirically we have observed that diagonal HMoGs consistently outperform their isotropic counterparts. We therefore focus on diagonal HMoGs throughout the remainder of this paper, noting that isotropic HMoGs offer negligible computational savings over diagonal ones.

\subsection{An efficient exponential family parameterization}\label{sec:complexity}

The HMoG model provides a unified framework for joint dimensionality reduction and clustering. However, its computational complexity becomes prohibitive for high-dimensional data or large numbers of latent components. To address this we analyze the complexity of of three HMoG architectures defined by constraints on the sufficient statistics $\V s_X$ and $\V s_Y$: \textit{(i)} a ``Full-Full HMoG'' where the sufficient statistics $\V s_X$ and $\V s_Y$ capture all second-order statistics within the observable and latent variables, respectively; \textit{(ii)} a ``Diagonal-Full HMoG'' where $\V s_X$ captures only diagonal second-order statistics while $\V s_Y$ captures full second-order statistics; and \textit{(iii)} a ``Diagonal-Diagonal HMoG'' where both $\V s_X$ and $\V s_Y$ are constrained to model only diagonal second-order statistics.

To begin, let us take the logarithm of Equation~\ref{eq:hmog-marginal} and express it as the difference between the data-dependent part
\begin{equation}\label{eq:posterior-part}
	\mathcal L_{data}(\V x) = \V s_X(\V x) \cdot \eprms_X + \psi_{YK}(\eprms_{YK} + \V x \cdot \iprms_{XY}) + \log\nu_X (\V x),
\end{equation}
and the data-independent part
\begin{equation}\label{eq:prior-part}
	\mathcal L_{model} = \psi_X(\eprms_X) + \psi_{YK}(\eprms_{YK} + \rprms_Y(\eprms_X, \iprms_{XY})).
\end{equation}
This separation is the core of our computational strategy, because we typically evaluate $\mathcal{L}_{data}$ much more frequently than $\mathcal L_{model}$ during inference, evaluation, and learning. In particular, we never require $\mathcal L_{model}$ during inference (Eq.~\ref{eq:hmog-posterior}), and when evaluating the log-likelihood $\sum_{i=1}^n \frac{1}{n} \log p(\V x^{(i)})$ (Eq.~\ref{eq:hmog-marginal}), $\mathcal{L}_{data}$ must be evaluated $n$ times, whereas $\mathcal{L}_{model}$ must be evaluated only once.

To continue, both $\mathcal L_{data}$ and $\mathcal L_{model}$ involve evaluating the log-partition function of the MoG $\psi_{YK}$. By inserting Equations~\ref{eq:categorical-log-partition} and~\ref{eq:mixture-conjugation} into Equation~\ref{eq:conjugated-log-partition}, we see that evaluating $\psi_{YK}$ reduces to $d_K$ evaluations of the latent MVN log-partition function $\psi_Y$. Therefore, depending on whether the inputs to $\psi_Y$ include full or diagonal precision matrices, the computational complexity of $\psi_{YK}$ is $\mathcal{O}(d_K \cdot d_Y^3)$ or $\mathcal{O}(d_K \cdot d_Y)$, respectively.

The distinct computations in $\mathcal L_{data}$ are the mapping $\V x \cdot \iprms_{XY}$ which has complexity $\mathcal O(d_X d_Y)$, and the term $\V s_X(\V x) \cdot \eprms_X$, which has complexity $\mathcal O(d_X^2)$ or $\mathcal O(d_X)$ depending on whether $\V s_X$ captures full or diagonal second-order statistics. For $\mathcal L_{model}$, the distinct computations are the term $\psi_X(\eprms_X)$, which has complexity $\mathcal O(d_X^3)$ or $\mathcal O(d_X)$, and the evaluation of the conjugation parameters $\rprms_Y$, which has complexity $\mathcal O(d_X^2 d_Y + d_X d_Y^2)$ or $\mathcal O(d_X d^2_Y)$ (Eqs.~\ref{eq:lgm-conjugation-mean}~\&~\ref{eq:lgm-conjugation-cov}). Critically however, because $\rprms_Y$ is always dense, $\psi_{YK}$ in $\mathcal L_{model}$ is always evaluated on full precision matrices.  We summarize the resulting complexities of the three model architectures in Table~\ref{tab:complexity}, and see how Diagonal-Diagonal HMoGs are unique in that $\mathcal L_{data}$ is much cheaper to evaluate than $\mathcal L_{model}$.

\begin{table}[t]
	\caption{\textit{The computational complexity of HMoGs.} Big-$\mathcal{O}$ complexity of evaluating the log-likelihood components for different HMoG parameterizations. Key variables are the observable dimension $d_X$, the latent dimension $d_Y$, and the number of components $d_K$.}
	\label{tab:complexity}

	\begin{center}
		\setlength{\tabcolsep}{1.5em} 
		\begin{tabular*}{\textwidth}{@{\extracolsep{\fill}}lcc@{}}
			\toprule
			\textbf{Architecture} & \multicolumn{2}{c}{\textbf{Complexity $\mathcal{O}(\cdot)$}} \\
			\cmidrule(lr){2-3}
			& $\mathcal{L}_{\text{data}}$ & $\mathcal{L}_{\text{model}}$ \\
			\midrule
			Full-Full & $d_X^2 + d_X d_Y + d_K d_Y^3$ & $d_X^3 + d_X^2 d_Y + d_X d_Y^2 + d_K d_Y^3$   \\
			Diagonal-Full & $d_X d_Y + d_K d_Y^3$ & $d_X d_Y^2 + d_K d_Y^3$ \\
			Diagonal-Diagonal & $d_X d_Y + d_K d_Y$ & $d_X d_Y^2 + d_K d_Y^3$ \\
			\bottomrule
		\end{tabular*}
	\end{center}
\end{table}

We conclude this section by analyzing the computational complexity of HMoG training (for algorithm details see~\cite{sokoloski_unified_2024}). For batch gradient ascent of the log-likelihood with batch size $b$, the per-batch computational cost is approximately $b \cdot \mathcal{O}(\mathcal{L}_{data}) + \mathcal{O}(\mathcal{L}_{model})$. For EM-based learning, the expectation step requires $n$ evaluations of $\mathcal{L}_{data}$. We implement the maximization step for HMoGs by taking $s$ gradient ascent steps per expectation step, each of which involves a single evaluation of $\mathcal{L}_{model}$. Since typically $n \gg s$, and for mini-batch learning $b > 1$, the computational cost of HMoG training is dominated by the repeated evaluation of $\mathcal{L}_{data}$. When training on large datasets, Diagonal-Diagonal HMoGs can therefore achieve orders of magnitude faster training times.

\section{Applications}

In this section we demonstrate HMoGs on MNIST, and show that joint optimization of dimensionality reduction and clustering improves performance over two-stage approaches. We focus our evaluation on model log-likelihood as a measure of fit, and Normalized Mutual Information (NMI) $\frac{2 \times I(K; C)}{H(K) + H(C)}$ as a measure of clustering performance, where $K$ are predicted cluster assignments, $C$ ground-truth classes, $I(K; C)$ is the mutual information, and $H(\cdot)$ is the entropy.

\subsection{Clustering MNIST}

We trained diagonal-diagonal HMoGs with $d_K = 10, 20, 40$, and $80$ clusters, and $d_Y = 10, 20, 50$, and $100$ dimensional-latent spaces on the MNIST training data. We trained 3 repeats of each configuration for 4000 iterations of gradient-ascent based EM (\cite{sokoloski_unified_2024}), using the Adam optimizer~\cite{kingma_adam_2014} with a learning rate of 1e-4, and 500 gradient steps per EM iteration. We also trained 3 repeats of each configuration with the two-stage approach (Sec.~\ref{sec:hmogs}). Deploying our JAX-based implementations on A100 GPUs, the run-time of a single fit of joint EM ranged from about 20 minutes to 2 hours. Two-stage fits were considerably faster as we used exact EM, and fit diagonal MoGs in the latent space, and thus avoided all dense-matrix inversions.

\begin{figure}[t]

	\begin{center}
		\includegraphics{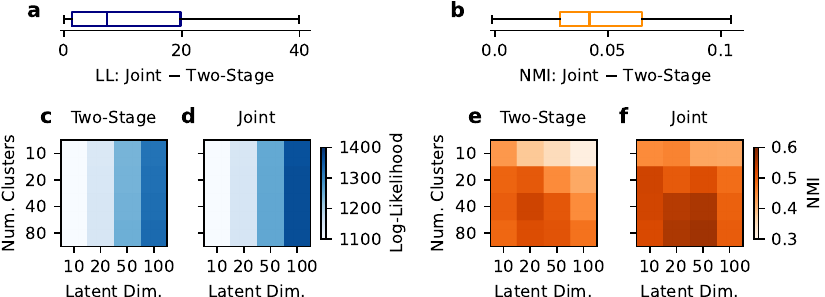}

		\vspace{6pt}

		\includegraphics[width=0.88\textwidth]{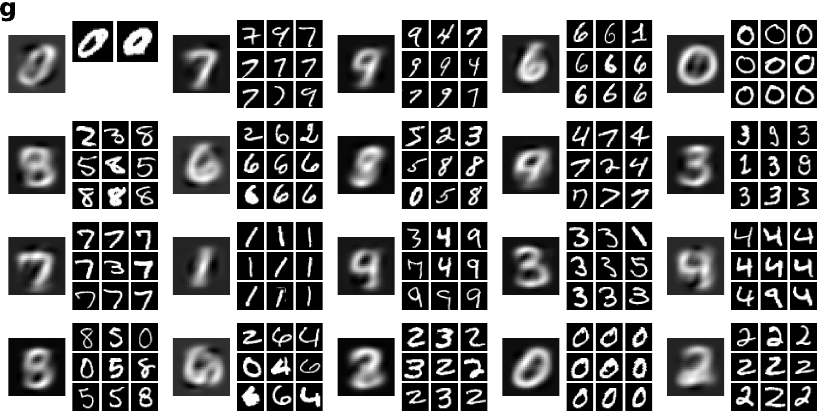}

	\end{center}

	\caption{\textit{Model evaluation of HMoGs versus two-stage models.} \textbf{a-b:} Distribution (boxplots) of average difference between joint and two-stage fits in log-likelihood (\textbf{a}) and NMI (\textbf{b}) over different configurations. \textbf{c-f:} Log-likelihood (\textbf{c-d}) and NMI (\textbf{e-f}) heatmaps for two-stage (\textbf{c, e}) and joint fits (\textbf{d, f}) across different latent dimensions and cluster numbers. Cluster prototypes (\textbf{g}, larger images) and example cluster members (smaller images) from a jointly trained model with $d_Y = 50$ and $d_K = 20$.
	}\label{fig:model-comparison}
\end{figure}

We find that joint EM consistently outperforms two-stage training across all configurations measured in both log-likelihood (Fig.~\ref{fig:model-comparison}\textbf{a}) and NMI (Fig.~\ref{fig:model-comparison}\textbf{b}) on the MNIST test data, which becomes more pronounced for more complex models. Log-likelihood performance generally improves with increasing latent dimension (Fig.~\ref{fig:model-comparison}\textbf{c-d}), whereas NMI performance tends to peak with intermediate latent dimension (Fig.~\ref{fig:model-comparison}\textbf{e-f}). Moreover, two-stage NMI peaks at intermediate cluster numbers (Fig~.\ref{fig:model-comparison}\textbf{f}), whereas joint optimization achieves peak NMI at the largest values tested. If we analyze cluster prototypes (expected observation $\E[X \mid K=i]$), we see that overclustering facilitates performance by allowing HMoGs to represent style variations within each digit class with multiple clusters (Fig.~\ref{fig:model-comparison}\textbf{g}).

%
%
%
\subsection{Sparse-coding and cluster merging}

We next develop an unsupervised cluster merging strategy to further improve NMI scores, and directly evaluate unsupervised classification accuracy. A key advantage of our probabilistic framework is principled uncertainty quantification, which we leverage to compute co-assignment probabilities $p(k=i|\V x^{(n)})p(k=j|\V x^{(m)})$ over all training images $\V x^{(i)}$, and thereby construct an average cluster similarity matrix. We filter out clusters with fewer then 30 cluster members, and use the resulting similarity matrix to compute an average-linkage hierarchical clustering (UPGMA) to determine final cluster associations.

\begin{figure}[t]

	\includegraphics{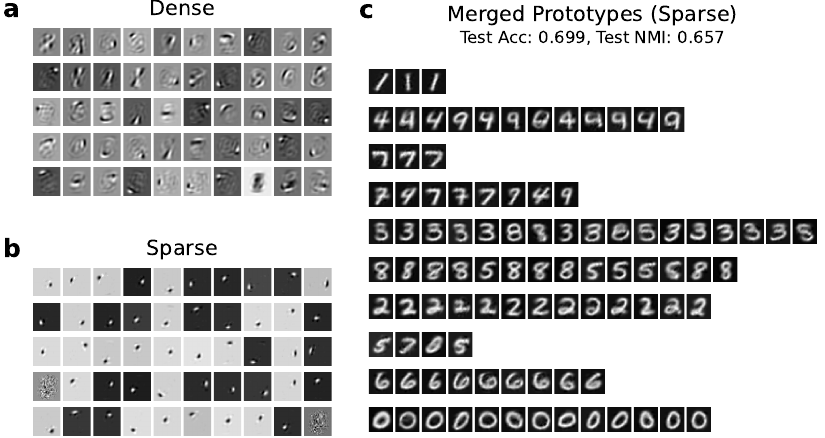}

	\caption{\textit{Effect of sparsity on loading matrices and clustering.} \textbf{a-b:} Loading vectors from an HMoG trained without (\textbf{a}) and with (\textbf{b}) $\ell_1$ sparsity. \textbf{c:} Performance metrics as well as cluster prototypes from the sparse HMoG, where each row contains all the  prototypes associated with one of ten classes based on co-assignment merging.}\label{fig:sparse-merge}\end{figure}

The capacity of HMoGs to efficiently support high-dimensional latent spaces also allows us to explore sparse-coding in our framework, where we regularize the HMoG to represent data with small subsets of latent features~\cite{vidal_sparse_2009}. Towards this end we trained 10 repeats of a diagonal-diagonal HMoG with $d_Y = 50$, $d_K=100$, and with sparsity $\ell_1=0$ or $\ell_1=0.01$, and in this case for 10,000 epochs of joint EM. By visualizing the columns of the loading matrix of two models, we see how unconstrained learning leads to holistic dimensionality reduction features (Fig.~\ref{fig:sparse-merge}\textbf{a}), whereas sparsity encourages highly localized, center-surround features (Fig.~\ref{fig:sparse-merge}\textbf{b}). We then merge learned clusters into 10 classes, and by visualizing cluster prototypes $\E[X \mid K=i]$ of our chose sparse model, we see that our merge strategy effectively isolates digit classes amongst learned clusters (Fig.~\ref{fig:sparse-merge}\textbf{c}).

Sparsity significantly improves the performance of our merging strategy. While sparse and dense HMoGs achieve similar pre‑merge clustering performance (mean NMI $0.542 \pm 0.009$ vs.\ $0.553 \pm 0.007$), sparse models gain significantly more from merging ($\Delta\mathrm{NMI} = 0.115 \pm 0.010$ vs.\ $0.046 \pm 0.016$, $p<0.001$), yielding higher final NMIs ($0.657 \pm 0.015$ vs.\ $0.600 \pm 0.053$, $p<0.01$). Sparse HMoGs also achieve better classification accuracy ($0.655 \pm 0.026$ vs.\ $0.532 \pm 0.032$, $p<0.0001$), Finally, sparse HMoGs exhibit $3.5$ times lower NMI variance across runs (SD $0.015$ vs.\ $0.053$), suggesting that sparsity also drives a more stable approach to clustering.

\section{Conclusion}

HMoGs bridge the gap between classical statistical rigor and modern machine learning by unifying dimensionality reduction and clustering into a single, tractable, probabilistic framework. Where similar probabilistic models struggle, HMoGs scale efficiently to hundreds of latent dimensions while maintaining exact inference, evaluation, and learning. This scalability of the latent space allows us to enforce sparse-coding in our probabilistic framework, so that data is encoded by a small number of features in the high-dimensional latent space.

Although sparse and non-sparse HMoGs achieve similar baseline clustering performance, sparse models benefit significantly more from our probabilistic cluster merging algorithm. We hypothesize that sparse codes sharpen cluster boundaries by driving non-overlapping clusters to be orthogonal in the latent space, and thereby facilitates higher-performance cluster merging. The emergence of center-surround features in sparse models also connects our work with neurobiologically-inspired coding principles~\cite{olshausen_emergence_1996}, and suggests sparse HMoGs may prove particularly useful when interpretable feature extraction is essential.

The clustering performance of HMoGs on MNIST (69.3\%) sits comfortably between classical k-means (53.5\%) and state-of-the-art deep methods (84.3\%)~\cite{xie_unsupervised_2016}. Nevertheless, the focus of our research was theoretical, and we believe HMoGs can achieve better performance through greater model complexity and improved training regimes. Moreover, their rigorous probabilistic formulation makes HMoGs well-suited as building blocks in more complex models. One avenue for driving state-of-the-art performance would be to use HMoGs to extend nonlinear probabilistic models that use MoGs as central components, such the normalizing flow-based GMMFlow~\cite{izmailov_semi-supervised_2020} or VAE-based VaDE architecture~\cite{jiang_variational_2017}. In future work, we look forward to pushing the frontiers of interpretable, model-based clustering through the continued development of hierarchical mixtures of Gaussians.

\newpage

\printbibliography

\end{document}